\documentclass[runningheads]{llncs}

\usepackage{amssymb}
\usepackage{amsmath}
\usepackage{algorithm}
\usepackage{algorithmic}
\usepackage{subcaption}
\usepackage{url}
\DeclareMathOperator*{\argmin}{arg\,min}

\usepackage{pgfplotstable}
\pgfplotsset{compat=1.8}
\usepgfplotslibrary{statistics}
\makeatletter
\pgfplotsset{
    boxplot prepared from table/.code={
        \def\tikz@plot@handler{\pgfplotsplothandlerboxplotprepared}%
        \pgfplotsset{
            /pgfplots/boxplot prepared from table/.cd,
            #1,
        }
    },
    /pgfplots/boxplot prepared from table/.cd,
        table/.code={\pgfplotstablecopy{#1}\to\boxplot@datatable},
        row/.initial=0,
        make style readable from table/.style={
            #1/.code={
                \pgfplotstablegetelem{\pgfkeysvalueof{/pgfplots/boxplot prepared from table/row}}{##1}\of\boxplot@datatable
                \pgfplotsset{boxplot/#1/.expand once={\pgfplotsretval}}
            }
        },
        make style readable from table=lower whisker,
        make style readable from table=upper whisker,
        make style readable from table=lower quartile,
        make style readable from table=upper quartile,
        make style readable from table=median,
        make style readable from table=lower notch,
        make style readable from table=upper notch
}
\makeatother

\pgfplotstableread{
min lq median uq max
2.52169 3.28163 3.57834 3.93879 5.59682
1.65469 2.05956 2.14348 2.20962 2.50632
2.30407 2.69645 2.74835 2.79946 3.42133
2.74791 3.05068 3.11868 3.19985 3.70411
2.96137 3.78047 3.87605 3.96503 4.38377
3.5516 3.92368 4.00042 4.08204 4.59311
3.29188 3.67751 3.75173 3.82773 4.19119
4.26828 4.65193 4.73502 4.82277 5.32202
3.39228 4.47612 4.59602 4.71269 5.20644
3.83426 4.44956 4.58876 4.73416 5.34594
3.38404 3.98575 4.13504 4.32864 5.13916
4.15886 5.07691 5.30238 5.54489 6.43874
2.83123 3.91162 4.07409 4.28907 4.97209
3.72364 4.86274 5.04945 5.1644 5.68575
4.22108 4.64694 4.70882 4.7799 5.07761
3.0198 4.20194 4.31614 4.42426 4.81512
2.68559 2.8484 2.91192 2.94902 3.0918
2.20827 2.26556 2.32084 2.34798 2.53656
0.632028 0.700401 0.745763 0.775047 1.00733
}\cifartrainnone

\pgfplotstableread{
min lq median uq max
2.5475 3.28589 3.58714 3.94489 5.4721
1.67928 2.06372 2.14516 2.21147 2.48877
2.38217 2.69633 2.74821 2.80115 3.33735
2.81122 3.05209 3.12129 3.20137 3.61017
2.93221 3.77879 3.87601 3.96342 4.40939
3.59801 3.92334 4.00085 4.0831 4.59806
3.29362 3.67708 3.75081 3.82732 4.13868
4.34524 4.65163 4.73482 4.82127 5.28756
3.69036 4.47328 4.59654 4.7106 5.26118
3.86148 4.44476 4.58368 4.73012 5.39171
3.45914 3.96988 4.12194 4.32272 5.16709
4.2961 5.03523 5.27955 5.53722 6.41517
2.73015 3.82935 4.02504 4.23867 4.99853
3.47208 4.78354 4.99388 5.13426 5.6276
4.25957 4.63938 4.69898 4.77291 5.07724
3.06113 4.18597 4.30993 4.42658 5.16426
2.68889 2.85118 2.91219 2.95068 3.10132
2.2057 2.28961 2.32911 2.3653 2.57379
0.632601 0.722784 0.760212 0.81725 1.06442
}\cifartestnone

\pgfplotstableread{
min lq median uq max
0.677702 1.02273 1.14229 1.27767 1.98433
1.07535 1.56179 1.62464 1.67877 1.97424
1.2517 1.53586 1.56826 1.60264 1.95887
1.53695 1.69602 1.72788 1.76091 1.966
1.50031 1.71131 1.74332 1.77376 1.95688
1.56144 1.73534 1.766 1.79851 1.95974
1.55557 1.74956 1.78348 1.81774 1.98079
1.44177 1.61305 1.64837 1.68558 1.90585
1.11658 1.47229 1.54759 1.62574 1.99555
0.636813 1.15181 1.32923 1.52394 1.99527
0.881279 1.53776 1.63056 1.71793 1.99434
1.08527 1.58477 1.68933 1.7679 1.99853
1.03073 1.72257 1.79519 1.85026 2.00625
1.12211 1.74314 1.80069 1.85483 2.00315
1.23495 1.77204 1.82861 1.87338 2.00312
1.75902 1.79513 1.82462 1.84252 1.9387
1.78977 1.88346 1.89965 1.93575 1.99187
1.85915 1.88674 1.90135 1.92383 1.99747
1.39316 1.55465 1.57633 1.62991 1.87708
}\cifartrainmaxgain

\pgfplotstableread{
min lq median uq max
0.69029 1.02502 1.14527 1.28005 1.8844
1.1316 1.56258 1.62559 1.6809 1.93864
1.2849 1.53629 1.56834 1.60412 1.93309
1.5823 1.69679 1.72895 1.76172 1.9554
1.51389 1.71185 1.74314 1.77364 1.95894
1.59058 1.73488 1.76649 1.79958 1.94977
1.61323 1.74826 1.78273 1.81798 1.94061
1.45417 1.61338 1.64914 1.68602 1.88513
1.13179 1.46763 1.54387 1.62511 1.90829
0.646588 1.09713 1.30664 1.51236 1.99214
0.77531 1.47084 1.60784 1.70586 1.98885
1.04508 1.52691 1.67067 1.75977 1.99983
0.808782 1.6466 1.78039 1.8398 1.99795
0.838761 1.68877 1.78969 1.84657 1.99778
1.0238 1.72217 1.8041 1.86533 2.00058
1.76105 1.79847 1.8307 1.85009 1.95126
1.79144 1.88552 1.90063 1.93384 1.9931
1.86031 1.88794 1.90451 1.93322 1.99527
1.28614 1.55494 1.57718 1.63269 1.87755
}\cifartestmaxgain

\begin{document}
\title{MaxGain: Regularisation of Neural Networks by Constraining Activation Magnitudes}
\titlerunning{MaxGain Regularisation of Neural Networks}
\author{
	Henry Gouk\inst{1} \and
	Bernhard Pfahringer\inst{1} \and
	Eibe Frank\inst{1} \and
	Michael Cree\inst{2}
}
\institute{
	Department of Computer Science, University of Waikato\\
	\email{hgrg1@students.waikato.ac.nz, \{bernhard.pfahringer,eibe.frank\}@waikato.ac.nz}
	\and
	School of Engineering, University of Waikato\\
	\email{michael.cree@waikato.ac.nz}
}
\maketitle

\begin{abstract}
Effective regularisation of neural networks is essential to combat overfitting due to the large number of parameters involved. We present an empirical analogue to the Lipschitz constant of a feed-forward neural network, which we refer to as the maximum gain. We hypothesise that constraining the gain of a network will have a regularising effect, similar to how constraining the Lipschitz constant of a network has been shown to improve generalisation. A simple algorithm is provided that involves rescaling the weight matrix of each layer after each parameter update. We conduct a series of studies on common benchmark datasets, and also a novel dataset that we introduce to enable easier significance testing for experiments using convolutional networks. Performance on these datasets compares favourably with other common regularisation techniques.
\end{abstract}

\section{Introduction}
%Spiel about regularisation being useful
% Reference some other Lipschitz related methods. They all assume the domain is R^d, but could perhaps be a low dimensional manifold in a high dimensional space?
% Introduce concept of an "empirical" Lipschitz constant, with the motivation being to constrain the Lipschitz constant for the manifold of interest, as inferred from the training data.
Regularisation is a crucial component in machine learning systems. This is particularly true for neural networks, whose huge number of parameters can lead to extreme overfitting, such as memorising the training set---even in the case where the labels have been randomised~\cite{zhang2016}. In this work, we investigate a regularisation technique inspired by recent work regarding the Lipschitz continuity of neural networks~\cite{gouk2018}. Most work in machine learning that deals with the concept of Lipschitz continuity assumes, often implicitly~\cite{gouk2018,miyato2018}, that the input domain of the function of interest is $\mathbb{R}^d$---sometimes with the additional assumption that each component in this vector space is bounded in, for example, the range $\lbrack -1, 1 \rbrack$. However, when working with unstructured data---a task at which neural networks excel---a common assumption is that the data lie in a low dimensional manifold embedded in a high dimensional space. This is known as the manifold hypothesis~\cite{cayton2005}. In this paper, we explore the idea of constraining the Lipschitz continuity of neural network models when they are viewed in this light: as mappings from the subset of $\mathbb{R}^d$ that contains the low dimensional manifold, to some meaningful vector space, such as the distribution over possible classes. The precise structure of the manifold is unknown to us, which makes constraining a function that operates on this manifold difficult. To circumvent this problem, we introduce the concept of \emph{gain}---an empirical analogue to the operator norm technique used by Gouk et al.~\cite{gouk2018} to compute the Lipschitz constant of a neural network layer.

We present a regularisation scheme that improves the generalisation performance of neural networks by constraining the maximum gain of each layer. This is accomplished using a simple modification to conventional neural network optimisers that applies a stochastic projection function in addition to a stochastic estimate of the gradient. We demonstrate the effectiveness of our regularisation algorithm on several classification datasets. A novel dataset that facilitates significance testing for convolutional network-based classifiers is introduced as part of these experiments. Additionally, we show how our technique performs when used in conjunction with other regularisation methods such as dropout~\cite{srivastava2014} and batch normalisation~\cite{ioffe2015}. We also provide empirical evidence that constraining the gain on the training set results in observing lower gain on the test set compared to when the gain on the training set is not constrained. Details of how the performance of models trained with out regularisation technique as its hyperparameter is varied are also provided.

\section{Related Work}
% Lipschitz related things: SN-GAN (other GANs, too), spectral norm regularisation?, my work.
Several recent publications have addressed the idea of Lipschitz continuity of neural networks. Most of this work has been on generative adversarial networks (GANs)~\cite{goodfellow2014}. Wasserstein GANs~\cite{arjovsky2017} are the first GAN variant that require some way of enforcing Lipschitz continuity in order to converge. They accomplish this by clipping each weight whenever its absolute value exceeds some predefined threshold. While this will maintain Lipschitz continuity, the Lipschitz constant will not be known. An alternative to weight clipping is to penalise the norm of the gradient of the critic network~\cite{gulrajani2017}, which has been shown to improve the stability of training Wasserstein GANs. This technique for constraining Lipschitz continuity is similar to ours, in the sense that it uses an approximate measure of the Lipschitz constant on the training data. It is, however, quite different in the sense that it is not being used for regularisation and that it is applied as a soft constraint using a penalty term. Miyato et al.~\cite{miyato2018} have also proposed normalising the weights in each layer of the discriminator network of a GAN using the spectral norm of the respective weight matrix, but they provide no evidence showing that their heuristic for applying this to convolutional layers actually constrains the spectral norm. Some recent work has shown how to precisely compute and constrain the Lipschitz constant of a network with respect to the $\ell_1$ and $\ell_\infty$ norms~\cite{gouk2018} and demonstrated that constraining the Lipschitz constant with respect to these norms has a regularising effect comparable to dropout and batch normalisation.

The idea of constraining the Lipschitz constant of a network is conceptually related to quantifying the flatness of minima. While there is no single formalisation for what constitutes a flat minimum, the unifying intuition is that a minimum is flat when a small perturbation of the model parameters does not have a large impact on the performance of the model. Dinh et al.~\cite{dinh2017} have shown that Lipschitz continuity is not a reliable tool for quantifying the flatness of minima. However, there is a subtle but very important difference between how they employ Lipschitz continuity, and how it is used by Gouk et al.~\cite{gouk2018} and in this work. Neural networks are functions parameterised by two distinct sets of variables: the model parameters, and the features. Dinh et al.~\cite{dinh2017} consider Lipschitz continuity with respect to the model parameters, whereas we consider Lipschitz continuity with respect the features being supplied to the network. The crux of the argument given by Dinh et al. is that the Lipschitz constant of a network with respect to its weights is not invariant to reparameterisation.

% Other regularisers: dropout (+drop connect), maxnorm, batchnorm, sparsity terms?, papers talking about how these interact with each other?
%Dropout~\cite{srivastava2014}, dropconnect~\cite{wan2013}, batchnorm~\cite{ioffe2015}
Dropout~\cite{srivastava2014} is one of the most widely used methods for regularising neural networks. It is popular because it is efficient and easy to implement,  requiring only that each activation is set to zero with some probability, $p$, during training. An extension proposed by Srivastava et al.~\cite{srivastava2014}, known as maxnorm, is to constrain the maginitude of the weight vector associated with each unit in some layer. One can also use multiplicative Gaussian noise, rather than Bernoulli noise. Kingma et al.~\cite{kingma2015} provide a technique that enables automatic tuning of the amount of noise that should be applied in the case of Gaussian dropout. A similar technique exists for automatically tuning $p$ for Bernoulli dropout---this extension is known as concrete dropout~\cite{gal2017}.

Batch normalisation~\cite{ioffe2015}, which was originally motivated by the desire to improve the convergence rate of neural network optimisers, is often used as a regularisation scheme. It is similar to our technique in the sense that it rescales the activations of a layer, but it does so in a different way: by standardising them and subsequenty multiplying them by a learned scale factor. Unlike other regularisation techniques, there is no hyperparameter for batch normalisation that can be tuned to control the capacity of the network. A similar technique, which does not rely on measuring activation statistics over minibatches, is weight normalisation~\cite{salimans2016}. This approach decouples the length and direction of the weight vector associated with each unit in the network, and enables one to train networks on very small batch sizes, which is a situation where batch normalisation cannot be applied reliably.

\section{Lipschitz Continuous Neural Networks}
\label{sec:lipschitz}
% Recap what was in the ICML submission
%	Lipschitz constant definition
%	Affine layer formalisation
%	Operator norm of the linear transform term is the Lipschitz constant
%	Easy to compute for L1 and Linf
Gouk et al. (2018)~\cite{gouk2018} recently demonstrated that constraining the Lipschitz constant of a neural network improves generalisation in the context of classification. We briefly review their technique to aid overall understanding and provide several useful definitions. Recall the definition of Lipschitz continuity:

\begin{equation}
\label{eq:lipscihtz}
D_B(f(\vec x_1), f(\vec x_2)) \leq k D_A(\vec x_1, \vec x_2) \quad \forall \vec x_1, \vec x_2 \in A,
\end{equation}

\noindent for some real-valued $k \geq 0$, and metrics $D_A$ and $D_B$. We refer to $f$ as being $k$-Lipschitz. We are most interested in the smallest possible value of $k$, which is sometimes referred to as the best Lipschitz constant. A particularly useful property of Lipschitz continuity is that the composition of a $k_1$-Lipschitz function with a $k_2$-Lipschitz function is a $k_1k_2$-Lipschitz function. Given that a feed-forward neural network can be expressed as a series of function compositions,

\begin{equation}
\label{eq:feed-forward}
f(\vec x) = (\phi_l \circ \phi_{l-1} \circ ... \circ \phi_1)(\vec x),
\end{equation}

\noindent one can compute the Lipschitz constant of the entire network by computing the constant of each layer in isolation and taking the product of these constants:

\begin{equation}
\label{eq:lipschitz-product}
L(f) = \prod_{i = 1}^{l} L(\phi_i),
\end{equation}

\noindent where $L(\phi_i)$ indicates the Lipschitz constant of some function, $\phi_i$.

Many functions in this product, such as commonly used activation functions and pooling operations, have a Lipschitz constant of one for all vector $p$-norms on $\mathbb{R}^d$. Other commonly used functions, such as fully connected and convolutional layers, can be expressed as affine transformations,

\begin{equation}
\label{eq:affine}
f(\vec x) = W \vec x + \vec b,
\end{equation}

\noindent where $W$ is a weight matrix and $\vec b$ is a bias vector. For fully connected layers, there is no special structure to $W$. In the case of convolutional layers, $W$ is a block matrix where each block is in turn a doubly block circulant matrix. Batch normalisation layers can also be expressed as affine transformations, where the linear operation is a diagonal matrix. Each element on the diagonal is one of the scaling parameters divided by the standard deviation of the corresponding activation. The Lipschitz constant of an affine function is given by the operator norm of the weight matrix,

\begin{equation}
\label{eq:operator-norm}
\|W\|_p = \sup_{\vec x \neq 0} \frac{\|W \vec x\|_p}{\|\vec x\|_p},
\end{equation}

\noindent for some vector $p$-norm. For the $\ell_1$ and $\ell_\infty$ vector norms, the matrix operator norms are given by the maximum absolute column sum and maximum absolute row sum norms, respectively. In the case of the $\ell_2$ norm, the operator norm of a matrix is given by the spectral norm---the largest singular value. This can be approximated for fully connected layers relatively efficiently using the power iteration method. Once the operator norms have been computed, projected gradient methods can be used to constrain the Lipschitz constant of each layer to be less than a user specified value.

%\subsection{Spectral Norm of Convolutional Layers}
% Exact Lipschitz constant of convolutional layers w.r.t. L2 norm
%	Use power method, but replace matmuls with convs
%There seems to be come confusion in the deep learning community about how one should compute the spectral norm of a convolutional layer. For example, [SN-GAN PAPER] reshape the tensor of convolution kernels into a matrix that does not correspond to the linear transformation used in a convolutional network and compute the spectral norm of this matrix. Gouk et al. (2018) simply state that they cannot see an obvious way to compute the spectral norm of a convolutional layer and instead focus their analysis on the $\ell_1$ and $\ell_\infty$ operator norms.

%It turns out that a rather small modification to the power iteration method can be used to compute the Lipschitz constant of a convolutional layer with respect to the $\ell_2$ vector norm. 

\section{Regularisation by Constraining Gain}
% Talk about the manifold assumption
A common assumption in machine learning is that many types of unstructured data, such as images and audio, lie near a low dimensional manifold embedded in a high dimensional vector space. This is known as the manifold hypothesis. If we assume that the manifold hypothesis holds, then a network will only be supplied with elements of some set $\mathcal{X} \subset \mathbb{R}^d$. As a consequence, the training procedure need only ensure that the network is Lipschitz continuous on $\mathcal{X}$ in order to construct a network with a slowly varying decision boundary. In practice, the exact structure of $\mathcal{X}$ is unknown, but we do have a finite sample of instances, $X \subset \mathcal{X}$, which we can use to empirically estimate various characteristics of $\mathcal{X}$. The work presented in this paper differs from that of Gouk et al. (2018)~\cite{gouk2018} in that an approximation of the Lipschitz constant on $\mathcal{X}$ is constrained, rather than the true Lipschitz constant on $\mathbb{R}^d$. The technique used to enforce the constraint during is detailed in this section.

\subsection{Gain}
% Show how the measured gain of the linear transform in some layers can be used to compute an empirical Lipschitz constant. Point out that labels are not required for estimating the constant.
Lipschitz continuity is not something that can be established empirically. However, one can find a lower bound for $k$ by sampling pairs of points from the training set and determining the smallest value of $k$ that satisfies Equation~\ref{eq:lipscihtz}. This solution, while conceptually simple, has a number of finer details that can greatly impact the result. For example, how should pairs be sampled? If they are chosen randomly, then a very large number of pairs will be required to provide a good estimate of $k$. On the other hand, if a hard-negative mining approach were employed, fewer pairs would be required, but the amount of computation per pair would be greatly increased.

By restricting our analysis to feed-forward neural networks, we derive a simpler and more computationally efficient approach. Recall that the Lipschitz constant of a feed-forward network is given by the product of the Lipschitz constants associated with each activation function---which are usually less than or equal to one and cannot be changed during training---and the operator norms associated with the linear transformations in the learned layers. We define gain using the fraction from Equation~\ref{eq:operator-norm},

\begin{equation}
\label{eq:gain}
Gain_p(W, \vec x) = \frac{\|W \vec x\|_p}{\|\vec x\|_p},
\end{equation}

%When X = \mathcal{X}, this is equivalent to the Lipschitz constant on \mathcal{X}
%Don't have to explicitly construct W
%Can use any p-norm---even those that are NP-hard to compute the induced operator norm.
\noindent for some input instance $\vec x$, and we use the maximum gain observed over some set of input vectors from our manifold of interest as an approximation of the operator norm. This empirical estimate of the operator norm of a matrix has several advantages over computing the true operator norm. Firstly, it fulfills our desire to approximately compute the Lipschitz constant of an affine function on $\mathcal{X}$. It is also well behaved, in the sense that $X = \mathcal{X} \implies \sup_{\vec x} Gain(W, \vec x) = \|W\|_p$. Some more practical advantages include not having to explicitly construct $W$, but merely requiring a means of computing $W \vec x$---a property that is extremely useful when computing the operator norm of a convolutional layer. Also, because one need not compute a matrix norm directly, it is possible to compute the gain with respect to a $p$-norm for which it would be NP-hard to compute the induced matrix operator norm.

\subsection{MaxGain Regularisation}
% Go over how this can be used for regularisation.
%	Limit the gain on a per-layer basis---if we are using the gain as an approxmation to the Lipschitz constant on \mathcal{X}, then \gamma^L will be an approximation of the Lipschitz constant of the whole network
%	Not feasible to iterate over the whole training set to compute the projection for a single batch
%	Pseudocode
The crux of our regularisation technique is to limit the gain of each layer in a feed-forward neural network. Each layer is constrained, in isolation, to have a gain less than or equal to a user specified hyperparameter, $\gamma$. Put formally, we wish to solve the following optimisation problem:

\begin{align}
\label{eq:objective}
W_{1 .. l} &= \argmin_{W_{1 .. l}} \sum_{\vec x_i^1 \in X} L(\vec x_i^1, \vec y_i) \\
\label{eq:constraint}
		   & s.t. \max_{\vec x_i^j} Gain_p(W_j, \vec x_i^j) \leq \gamma \qquad \forall j \in \{1 \, ... \, l\},
\end{align}

\noindent where $\vec x_i^j$ indicates the input to the $j$th layer for instance $i$, $\vec y_i$ is a label vector associated with instance $i$, $W_j$ is the weight matrix for layer $j$, and $L(\cdot)$ is some task-specific loss function. Note that if $\|\vec x_i^j\|_p$ is zero, we set the gain for that particular measurement to zero rather than leaving it undefined.

The conventional approach to solving Equation~\ref{eq:objective} without the constraint in Equation~\ref{eq:constraint} is to use some variant of the stochastic gradient method. For simple constraints, such as requiring $W_j$ to lie in some known convex set, a projection function can be used to enforce the constraint after each parameter update. In our case, applying the projection function after each parameter update would involve propagating the entire training set through the network to measure the maximum gain for each layer. Even for modest sized datasets this is completely infeasible, and it defeats the purpose of using a stochastic optimiser. Instead, we propose the use of a stochastic projection function, where the $\max$ in Equation~\ref{eq:constraint} is taken over the same minibatch used to compute an estimate of the loss function gradient. We reuse the ``stale'' activations computed before the weight update in order to avoid the extra computation required for propagating all of the instances through the network again. The following projection function is used:

\begin{equation}
\label{eq:proj}
\pi(W, \hat{\gamma}, \gamma) = \frac{1}{\max(1, \frac{\hat{\gamma}}{\gamma})} W,
\end{equation}

\noindent where $\hat{\gamma}$ is our estimate of the maximum gain for layer $j$. If the MaxGain constraint is not violated, then $W$ will be left untouched. If the constraint is violated, $W$ will be rescaled to fix the violation. In the case where the maximum gain is computed exactly, this function will rescale the weight matrix such that the maximum gain is less than or equal to $\gamma$. Because we are only approximately computing the maximum gain, this constraint will not be perfectly satisfied on the training set.

During training, batch normalisation applies a transformation to the activations of a minibatch using statistics computed using only the instances contained in that minibatch. Thus, the gain measured for a particular instance is dependent on the other instances in the batch in which it is observed by the network. Specifically, the activations, $\vec x$, produced by some layer, are standardised:

\begin{equation}
\phi^{bn}(\vec x) = \text{diag}(\frac{\vec \alpha}{\sqrt{\text{Var}\lbrack \vec x \rbrack}}) (\vec x - \text{E}\lbrack \vec x \rbrack) + \vec \beta,
\end{equation}

\noindent where $\text{diag}(\cdot)$ denotes a diagonal matrix, $\vec \alpha$ and $\vec \beta$ are learned parameters, and the $\text{Var}\lbrack \cdot \rbrack$ and $\text{E}\lbrack \cdot \rbrack$ operations are computed over only the instances in the current minibatch. If the estimated mean and variance values are particularly unstable, then the gain values will also be very unstable and the training procedure will converge very slowly---or possibly not at all. We have found that the high dimensionality of neural network hidden layer activation vectors, and their sparse nature when using the ReLU activation function, coupled with a relatively small batch size, leads to unstable measurements when using MaxGain in conjunction with batch normalisation. We remedy this by recomputing the batch normalisation output in the projection function using the running averages of the standard deviation estimates that are kept for performing test-time predictions. By standardising the minibatch activations using these more stable estimates of the activation statistics, we observed considerably more reliable convergence. Note that the stochastic estimates of the mean and standard deviation of activations are still used for computing the gradient---it is only the projection function that uses the running averages of these values.

Pseudocode for our constrained optimisation algorithm based on stochastic projection is provided in Algorithm~\ref{alg:pseudocode}. The inputs to each layer for each minibatch, $X_{1:l}^{(t)}$, and the results of transforming these by the linear term of the affine transformations, $Z_{1:l}^{(t)}$, are cached during the gradient computation to be reused in the projection function. We use a single hyperparameter, $\gamma$, to control the allowed gain of each layer. There is no fundamental reason that a different $\gamma$ cannot be selected for each layer other than the added difficulty in optimising more hyperparameters. The $update(\cdot, \cdot)$ function can be any stochastic optimisation algorithm commonly used with neural networks. We consider both Adam~\cite{kingma2014} and SGD with Nesterov momentum.

\begin{algorithm}
\caption{This algorithm makes use of the stochastic gradient method (or some variation thereof) and a stochastic projection function to approximately solve the constrained optimisation problem outlined in Equations~\ref{eq:objective} and \ref{eq:constraint}. In this procedure, $zip(\cdot, \cdot)$ constructs an array of tuples consisting of elements from the sequences passed as arguments.}
\label{alg:pseudocode}
\begin{algorithmic}
%\Require{$f(W_{1:l}^{(0)})$, a network with initial parameters $W_{1}^{(0)}, W_{2}^{(0)}, \hdots, W_{l}^{(0)}$}
%\Require{$\lambda$, the desired upper bound for each $\|W_i\|_p$}
\STATE $t \gets 0$
\WHILE{$W_{1:l}^{(t)}$ not converged}
	\STATE $t \gets t + 1$
	\STATE $(g_{1:l}^{(t)}, X_{1:l}^{(t)}, Z_{1:l}^{(t)}) \gets \nabla_{W_{1:l}} f(W_{1:l}^{(t-1)})$
	\STATE $\widehat{W}_{1:l}^{(t)} \gets update(W_{1:l}^{(t-1)}, g_{1:l}^{(t)})$
	\FOR{$i = 1$ \textbf{to} $l$}
    	\STATE $\hat{\gamma} \gets 0$
        \FOR{$(\vec x_j, W_i^{(t)}\vec x_j)$ \textbf{in} $zip(X_i^{(t)}, Z_i^{(t)})$}
        	\STATE $\hat{\gamma} \gets max(\hat{\gamma}, \frac{\|W_i^{(t)}\vec x_j\|_p}{\|x_j\|_p})$
        \ENDFOR
		\STATE $W_{i}^{(t)} \gets \pi(\widehat{W}_{i}^{(t)}, \hat{\gamma}, \gamma)$
	\ENDFOR
\ENDWHILE
\end{algorithmic}
\end{algorithm}

\subsection{Compatibility with Dropout}
% Mention details about dropout that were discussed in more detail in the ICML submission.
There are two parts to applying dropout regularisation to a network. Firstly, during training, one must stochastically corrupt the activations of some hidden layers, usually by multiplying them with vectors of Bernoulli random variables. Secondly, during test time, the activations are scaled such that the expected magnitude of each activation is the same as what it would have been during training. In the case of standard Bernoulli dropout, this just means multiplying each activation by the probability that it was not corrupted during training. This scaling is known to change the Lipschitz constant of a network over $\mathbb{R}^d$~\cite{gouk2018}, and the same argument applies to the Lipschitz constant on $\mathcal{X}$. Because many commonly used activation functions are homogeneous, namely ReLU and its many variants, scaling the output activations is equivalent to scaling the output of the affine transformation. This, in turn, has an identical effect to scaling both the weight matrix and bias vector. Due to the homogeneity of norms, this scaling also directly affects the gain. Therefore, one might expect that one needs to increase $\gamma$ when using our technique in conjunction with dropout.

\section{Experiments}
The experiments reported in this section aim to demonstrate several aspects of our MaxGain regularisation method. The primary question we wish to answer is whether our technique for constraining the maximum gain of each learned layer in a network is an effective regularisation method. We also demonstrate that constraining the gain on training instances results in observing lower gain on the test, compared to when the gain is not constrained at all. All networks trained with MaxGain regularisation use the same $\gamma$ parameter for each layer in order to simplify hyperparameter optimisation. While the method we have presented can be used in conjunction with any vector norm, in this work we only investigate how well MaxGain works when using the $\ell_2$ vector norm.

Throughout our experiments, we make use of several different datasets. We also introduce a novel dataset larger than some typical benchmark datasets, like CIFAR-10 and MNIST, yet smaller and more manageable than the ImageNet releases used for the Large Scale Visual Recognition challenges. This dataset is designed so that performing significance tests is easy, and a greater degree of confidence can therefore be attributed to conclusions drawn from experiments using this dataset. The pixel intensities of all images have been scaled to lie in the range $\lbrack -1, 1 \rbrack$.

\subsection{CIFAR-10}
CIFAR-10~\cite{krizhevsky2009} is a collection of 60,000 tiny colour images, each labelled with one of 10 classes. In our experiments we follow the standard protocol of using 50,000 images for training and 10,000 images for testing. Additionally, we use a 10,000 image subset of the training set to tune the hyperparameters. We use the VGG-19 network~\cite{simonyan2014} trained using the Adam optimiser~\cite{kingma2014}. The model is trained for 140 epochs, starting with a learning rate of $10^{-4}$, which is decreased to $10^{-5}$ at epoch 100 and $10^{-6}$ at epoch 120. We make use of data augmentation in the form of horizontal flips, and padding training images to $40\times40$ pixels and cropping out a random $32\times32$ patch.

Results demonstrating how our technique compares with other common regularisation techniques are given in Table~\ref{tbl:cifar-10}. Several trends stand out in this table. Firstly, when comparing with each other technique in isolation, our method performs noticeably better than dropout and similarly to batch normalisation. When used in conjunction with batch normalisation the resulting test accuracy improves further. Interestingly, combining the use of dropout with both other regularisation approaches does not seem to have a noticeable cumulative effect.

\begin{table}
	\center
	\caption{Accuracy of a VGG-19 network trained in CIFAR-10 with different regularisation techniques.}
	\label{tbl:cifar-10}
	\begin{tabular}{lc}
		\hline
		Regulariser & Accuracy \\
		\hline
		None & 88.29\% \\
		Dropout & 89.71\% \\
 %       $\ell_1$-Lipschitz & 89.38\% \\
 %       $\ell_\infty$-Lipschitz & 88.90\% \\
		Batchnorm & 90.80\% \\
		Batchnorm + Dropout & 90.90\% \\
		\hline
		MaxGain & 90.75\% \\
		MaxGain + Dropout & 90.95\% \\
		MaxGain + Batchnorm & 91.76\% \\
		MaxGain + Batchnorm + Dropout & 91.52\% \\
		\hline
	\end{tabular}
\end{table}

\subsection{CIFAR-100}
CIFAR-100~\cite{krizhevsky2009} is similar to CIFAR-10, on account of containing 60,000 colour images of size $32 \times 32$, also split into a predefined set of $50,000$ for training and $10,000$ for testing. It differs in that it contains 100 classes, and exhibits more subtle inter-class variation. We use a Wide Residual Network~\cite{zagoruyko2016} on this dataset, in order to investigate how well MaxGain works on networks with residual connections. Batch normalisation is applied to all models trained on this dataset. We found convergence to be unreliable when training Wide ResNets without batch normalisation. Stochastic gradient descent with Nesterov momentum is used to train for a total of 200 epochs. We start with a learning rate of $10^{-1}$ and decrease by a factor of $5$ at epochs $60$, $120$, and $160$. We use the same data augmentation as was used for the CIFAR-10 models.

Results for experiments run on CIFAR-100 are given in Table~\ref{tbl:cifar-100}. In this case, we can see that our method performs comparably to dropout when both techniques are  used in conjunction with batch normalisation. The combination of all three regularisation schemes performs the best.

\begin{table}
	\center
	\caption{Accuracy of a Wide Residual Network with a depth of 16 and a width factor of four trained on CIFAR-100 with different regularisation techniques.}
	\label{tbl:cifar-100}
	\begin{tabular}{lc}
		\hline
		Regulariser & Accuracy \\
		\hline
		Batchnorm & 75.34\% \\
		Batchnorm + Dropout & 75.72\% \\
%        Batchnorm + $\ell_1$-Lipschitz & 76.12\% \\
%        Batchnorm + $\ell_\infty$-Lipschitz & 76.51\% \\
		\hline
		MaxGain + Batchnorm & 75.89\% \\
		MaxGain + Batchnorm + Dropout & 76.44\% \\
		\hline
	\end{tabular}
\end{table}

\subsection{Street View House Numbers (SVHN)}
The Street View House Numbers Dataset contains over 600,000 colour images depicting house numbers extracted from Google street view photos. Each image is $32\times32$ pixels, and the dataset has a predefined train and test split of 604,388 and 26,032 images, respectively. The distributions of the training and test splits are slightly different, in that the majority of the training images are considered less difficult. We train a VGG-style network on this dataset using the Adam~\cite{kingma2014} optimiser. Likely due to the large size of the dataset, we found that the network only needed to be trained for 17 epochs. We began with a learning rate of $10^{-4}$ and reduced it by a factor of 10 for the last two epochs.

Table~\ref{tbl:svhn} shows how the different models we considered performed on SVHN. An interesting result here is, in isolation, dropout outperforms both MaxGain and batch normalisation in terms of accuracy improvement over the baseline. This is potentially due to the mismatch between the distributions of the training and testing datasets. Despite the lackluster performance of these methods in isolation, they do still provide a benefit when combined with each other and dropout, which is consistent with the results of our other experiments.

\begin{table}
	\center
	\caption{Accuracy of a VGG-style network on the SVHN dataset when trained with various regularisation techniques.}
	\label{tbl:svhn}
	\begin{tabular}{lc}
		\hline
		Regulariser & Accuracy \\
		\hline
		None & 96.99\% \\
		Dropout & 97.72\% \\
%        $\ell_1$-Lipschitz & 97.64\% \\
%        $\ell_\infty$-Lipschitz & 97.42\% \\
		Batchnorm & 96.97\% \\
		Batchnorm + Dropout & 97.86\% \\
		\hline
		MaxGain & 97.22\% \\
		MaxGain + Dropout & 97.89\% \\
		MaxGain + Batchnorm & 97.31\% \\
		MaxGain + Batchnorm + Dropout & 97.98\% \\
		\hline
	\end{tabular}
\end{table}

\subsection{Scaled ImageNet Subset (SINS-10)}
Many datasets used by the deep learning community consist of a single predefined training and test split. For example, in the previous experiments on CIFAR-10 we stated that a set of 50,000 images was used for training, and another set of 10,000 images was used for testing. In order to perform some sort of significance test, and thus have some degree of confidence in our results and the conclusions we draw from them, we must gather multiple measurements of how well models trained using a particular algorithm configuration perform. To this end, we propose the Scaled ImageNet Subset (SINS-10) dataset, a set of 100,000 colour images retrieved from the ImageNet collection~\cite{deng2009}. The images are evenly divided into 10 different classes, and each of these classes is associated with multiple synsets from the ImageNet database. All images were first resized such that their smallest dimension was 96 pixels and their aspect ratio was maintained. Then, the central $96 \times 96$ pixel subwindow of the image was extracted to be used as the final instance. The labelled images are available online.\footnote{\url{https://www.cs.waikato.ac.nz/~ml/sins10/}}

An important difference between the proposed dataset and currently available benchmark datasets is how it has been split into training and testing data. The entire dataset is divided into 10 equal sized predefined folds of 10,000 instances. The first 9,000 images in each fold are intended for training a model, and the remaining 1,000 for testing it. One can then apply a machine learning technique to each fold in the dataset, and repeat the process for techniques one wishes to compare against. This will result in 10 performance measurements for each algorithm. A paired {\em t}-test can then be used to determine whether there is a significant difference, with some level of confidence, between the performance of the different techniques.

Note that the protocol for SINS-10 is different to the commonly used cross-validation technique. When performing cross-validation, the training sets overlap significantly, and the measurements for the test fold performance are therefore not independent. To mitigate this, one can use a heuristic for correcting the paired {\em t}-test~\cite{nadeau2000}. Rather than use this heuristic, we simply avoid fitting models using overlapping training (or test) sets, and can therefore use the standard paired {\em t}-test.

We train a Wide Residual Network with a width factor of four on this dataset. No data augmentation was used and each model was trained for 90 epochs using stochastic gradient descent with Nesterov momentum. The learning rate was started at $10^{-1}$ and decreased by a factor of five at epochs 60 and 80. For each regularisation scheme, we trained a model on each fold of the dataset. Regularisation hyperparameters, such as $\gamma$ and the dropout rate, were determined on a per-fold basis using a validation set of 1,000 instances drawn from the training set of the fold under consideration.

\begin{table}
\centering
\caption{Performance of the Wide Residual Network on the Scaled ImageNet Subset dataset using various combinations of regularisation techniques. The figures in this table are the mean accuracy $\pm$ the standard error, as measured across the 10 different folds.}
\label{tbl:sins}
\begin{tabular}{lc}
\hline
Regulariser & Accuracy \\
\hline
Batchnorm & $70.13\%$ $(\pm 0.27)$ \\
Batchnorm + Dropout & $74.81\%$ $(\pm 0.49)$ \\
%Batchnorm + $\ell_1$-Lipschitz & $71.43\%$ $(\pm 0.32)$ \\
%Batchnorm + $\ell_\infty$-Lipschitz & $72.49\%$ $(\pm 0.35)$ \\
\hline
MaxGain + Batchnorm & $70.65\%$ $(\pm 0.54)$ \\
MaxGain + Batchnorm + Dropout & $74.80\%$ $(\pm 0.51)$ \\
\hline
\end{tabular}
\end{table}

Results for the different regularisation schemes trained on this dataset are given in Table~\ref{tbl:sins}. We report the mean accuracy across each of the 10 folds, as well as the standard error. Paired {\em t}-tests were performed for comparing Batchnorm to MaxGain + Batchnorm, and also for Batchnorm + Dropout versus MaxGain + Batchnorm + Dropout. Neither of the tests resulted in a statistically significant difference ($p=0.332$ and $p=0.976$, respectively).

\subsection{Gain on the Test Set}
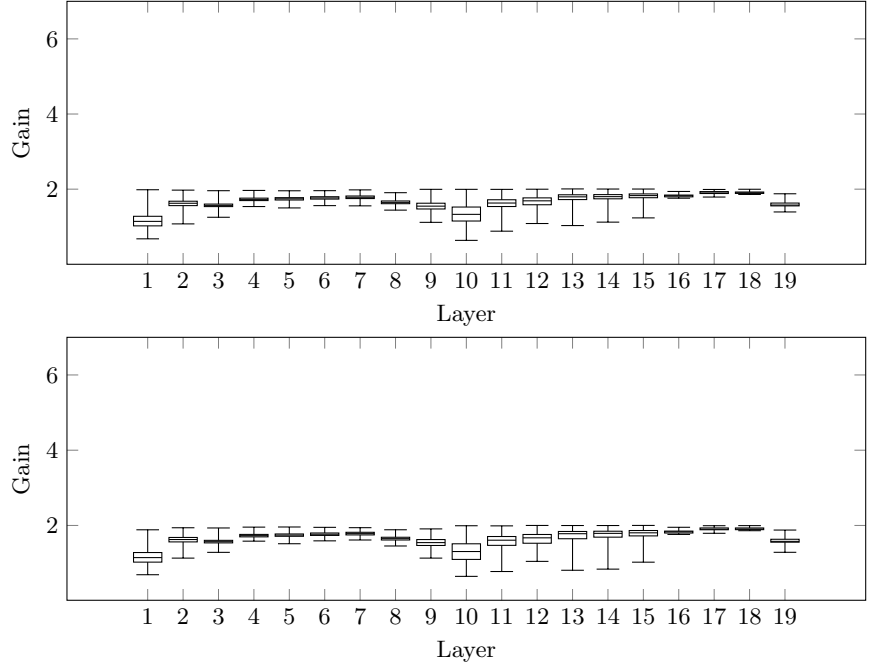
\begin{figure}
\centering
\begin{tikzpicture}
\begin{axis}[width=\textwidth,height=2in, boxplot/draw direction=y, xtick={1,2,...,19}, xticklabels={1,2,...,19}, ymax=7, ylabel={Gain}, xlabel={Layer}, ytick={2,4,6}]
\pgfplotstablegetrowsof{\cifartrainmaxgain}
\pgfmathtruncatemacro\TotalRows{\pgfplotsretval-1}
\pgfplotsinvokeforeach{0,...,\TotalRows}
{
  \addplot [
  boxplot prepared from table={
    table=\cifartrainmaxgain,
    row=#1,
    lower whisker=min,
    upper whisker=max,
    median=median,
    lower quartile=lq,
    upper quartile=uq
  },
  boxplot prepared
  ]
  coordinates {};

}
\end{axis}
\end{tikzpicture}
\begin{tikzpicture}
\begin{axis}[width=\textwidth,height=2in, boxplot/draw direction=y, xtick={1,2,...,19}, xticklabels={1,2,...,19}, ymax=7, ylabel={Gain}, xlabel={Layer}, ytick={2,4,6}]
\pgfplotstablegetrowsof{\cifartestmaxgain}
\pgfmathtruncatemacro\TotalRows{\pgfplotsretval-1}
\pgfplotsinvokeforeach{0,...,\TotalRows}
{
  \addplot [
  boxplot prepared from table={
    table=\cifartestmaxgain,
    row=#1,
    lower whisker=min,
    upper whisker=max,
    median=median,
    lower quartile=lq,
    upper quartile=uq
  },
  boxplot prepared
  ]
  coordinates {};

}
\end{axis}
\end{tikzpicture}
\caption{Boxplots showing the distributions of gains measured on each layer of the MaxGain-regularised VGG-19 network trained on CIFAR-10. The top plot shows the distributions on the training set, and the bottom plot on the test set.}
\label{fig:cifar10-gains-maxgain}
\end{figure}

Due to the stochastic nature of the projection function, the technique used to constrain the gain on the training set is only approximate. Therefore, it is important that we verify whether the constraint is fulfilled in practice. Moreover, even if the constraint is satisfied on the training set, that does not necessarily mean it will be satisfied on data not seen during training. To investigate this, we supply plots in Figure~\ref{fig:cifar10-gains-maxgain} showing the distribution of gains in each layer in the VGG-19 network trained using MaxGain on the CIFAR-10 dataset. We can see that the distributions between the train and test sets are virtually identical, and are never significantly above 2---the value selected for $\gamma$ when training this network.

\begin{figure}
\centering
\begin{tikzpicture}
\begin{axis}[width=\textwidth,height=2in, boxplot/draw direction=y, xtick={1,2,...,19}, xticklabels={1,2,...,19}, ylabel={Gain}, xlabel={Layer}]
\pgfplotstablegetrowsof{\cifartrainnone}
\pgfmathtruncatemacro\TotalRows{\pgfplotsretval-1}
\pgfplotsinvokeforeach{0,...,\TotalRows}
{
  \addplot [
  boxplot prepared from table={
    table=\cifartrainnone,
    row=#1,
    lower whisker=min,
    upper whisker=max,
    median=median,
    lower quartile=lq,
    upper quartile=uq
  },
  boxplot prepared
  ]
  coordinates {};

}
\end{axis}
\end{tikzpicture}
\begin{tikzpicture}
\begin{axis}[width=\textwidth,height=2in, boxplot/draw direction=y, xtick={1,2,...,19}, xticklabels={1,2,...,19}, ylabel={Gain}, xlabel={Layer}]
\pgfplotstablegetrowsof{\cifartestnone}
\pgfmathtruncatemacro\TotalRows{\pgfplotsretval-1}
\pgfplotsinvokeforeach{0,...,\TotalRows}
{
  \addplot [
  boxplot prepared from table={
    table=\cifartestnone,
    row=#1,
    lower whisker=min,
    upper whisker=max,
    median=median,
    lower quartile=lq,
    upper quartile=uq
  },
  boxplot prepared
  ]
  coordinates {};

}
\end{axis}
\end{tikzpicture}
\caption{Boxplots showing the distributions of gains measured on each layer of the unregularised VGG-19 network trained on CIFAR-10. The top plot shows the distributions on the training set, and the bottom plot on the test set.}
\label{fig:cifar10-gains-none}
\end{figure}
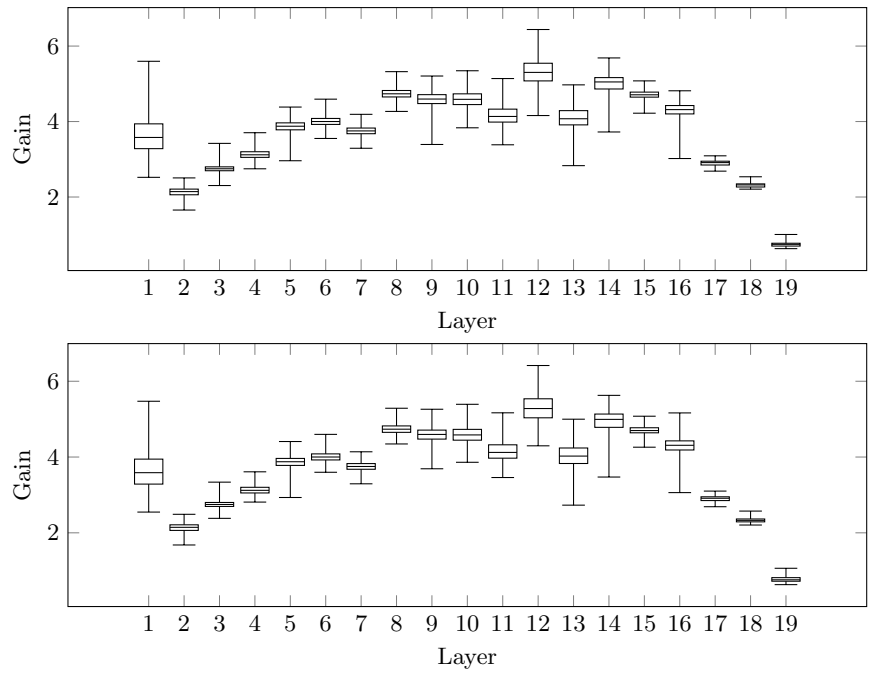

In addition to demonstrating that the stochastic projection function does effectively limit the maximum gain on the test set, we find it interesting to visualise gain measurements taken from each layer in a network trained without the MaxGain regulariser. This visualisation is given in Figure~\ref{fig:cifar10-gains-none}. Once again, the distributions of gains measured on the training versus test data are almost identical. Comparing the distributions given in Figure~\ref{fig:cifar10-gains-none} with those provided in Figure~\ref{fig:cifar10-gains-maxgain} show that the MaxGain regulariser has a substantial effect on the activation magnitudes produced by each layer.

If there is no constraint on the magnitude of the weights, then once the network can almost perfectly classify the training data, the optimiser can easily decrease the log loss by making the weights bigger. This results in an ``exploding activation'' effect, similar to the exploding/vanishing gradient phenomenon, which is only curbed when the cost of the small number of instances in the training set that are very confidently classified incorrectly begin to outweigh the increase in confidence on the correct classifications. Because MaxGain constrains the weight sizes of each layer, those that would have had large weights no longer do, and those that would have had small weights will now need larger weights in order to increase the confidence of the model. This results in the far more uniform changes in activation magnitude in Figure~\ref{fig:cifar10-gains-maxgain} compared to those in Figure~\ref{fig:cifar10-gains-none}.

\subsection{Sensitivity to $\gamma$}
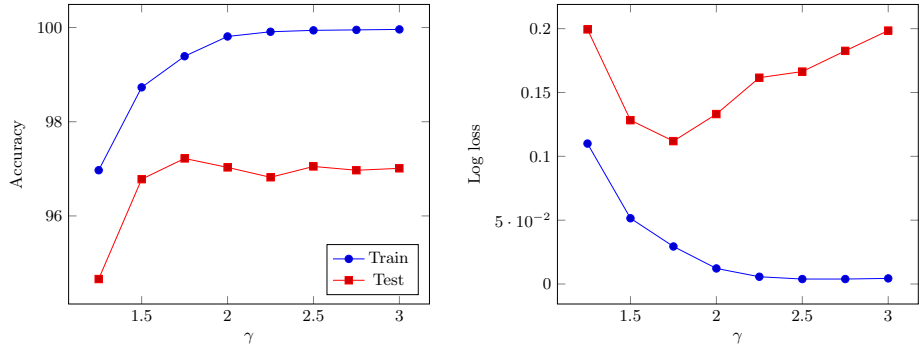
\begin{figure}
\centering

\begin{subfigure}[b]{0.49\textwidth}
\begin{tikzpicture}[scale=0.7]
\begin{axis}[xlabel=$\gamma$, ylabel=Accuracy, legend pos=south east]
\addplot coordinates {
	(1.25, 96.97)
    (1.5, 98.73)
    (1.75, 99.39)
    (2, 99.81)
    (2.25, 99.91)
    (2.5, 99.94)
    (2.75, 99.95)
    (3, 99.96)
};

\addplot coordinates {
	(1.25, 94.66)
    (1.5, 96.78)
    (1.75, 97.22)
    (2, 97.03)
    (2.25, 96.82)
    (2.5, 97.05)
    (2.75, 96.97)
    (3, 97.01)
};
\legend{{Train}, {Test}}
\end{axis}
\end{tikzpicture}
\end{subfigure}
\begin{subfigure}[b]{0.49\textwidth}
\begin{tikzpicture}[scale=0.7]
\begin{axis}[xlabel=$\gamma$, ylabel=Log loss]
\addplot coordinates {
	(1.25, 0.1100)
    (1.5, 0.0515)
    (1.75, 0.0294)
    (2, 0.0122)
    (2.25, 0.0057)
    (2.5, 0.0039)
    (2.75, 0.0039)
    (3, 0.0044)
};

\addplot coordinates {
	(1.25, 0.1994)
    (1.5, 0.1283)
    (1.75, 0.1119)
    (2, 0.1331)
    (2.25, 0.1616)
    (2.5, 0.1664)
    (2.75, 0.1826)
    (3, 0.1985)
};

\end{axis}
\end{tikzpicture}
\end{subfigure}
\caption{Accuracy (left) and log loss (right) of the VGG-style model on both the train and test splits of the SVHN dataset as the $\gamma$ hyperparameter is varied. The legend is shared between both plots.}
\label{fig:svhn-gamma}
\end{figure}

% Should behave similarly to constraining Lipschitz constant: small \gamma should underfit, large \gamma should lead to overfitting.
The single hyperparameter, $\gamma$, that is used to control the capacity of MaxGain-regularised networks should behave similarly to the $\lambda$ hyperparameter proposed by Gouk et al.~\cite{gouk2018} which is used to precisely bound the Lipschitz constant. In particular, when $\gamma$ is set to a small value the model should underfit, and when it is set to a large value one should observe overfitting. We explore this empirically in the context of the VGG-style network trained on SVHN. Figure~\ref{fig:svhn-gamma} shows how the performance on the training and test sets of SVHN varies as $\gamma$ is changed. This plot shows that $\gamma$ behaves in much the same way as the previously mentioned $\lambda$ hyperparameter. Specifically, for very low values of $\gamma$, the network exhibits low accuracy and high loss for both the train and test splits of the dataset. As the value of $\gamma$ is increased, the training accuracy goes towards 100\% and the loss goes towards zero. The test accuracy peaks and then plateaus, however the loss on the training set continues to increase, indicating that the network is more confidently misclassifying instances rather than misclassifying more instances.

\section{Conclusion}
This paper introduced MaxGain, a method for regularising neural networks by constraining how the magnitudes of activation vectors can vary across layers. It was shown how this method can be seen as an approximation to constraining the Lipschitz constant of a network, with the advantage of being usable for any vector norm. The technique is conceptually simple and easy to implement efficiently, thus making it a very practical approach to controlling the capacity of neural networks. We have shown that MaxGain performs competitively with other common regularisation schemes, such as batch normalisation and dropout, when compared in isolation. It was also demonstrated that when these techniques are combined together, further performance gains can be achieved. We envision that people will be most interested in using our method in conjunction with other regularisation schemes, as all of the methods used in our experiments add little runtime overhead to the training process.

Some of the results presented in this paper were obtained using a novel dataset with predefined folds that allows for practical significance testing in experiments involving convolutional networks. We hope that this moderately sized dataset will enable more confident conclusions to be drawn from future experiments.

\bibliography{refs}
\bibliographystyle{splncs04}
\end{document}